\documentclass{bioinfo}
\copyrightyear{2016} \pubyear{2016}

\access{ISMB 2016 \\ This is the Author's Original Version (AOV). This article has been published by Oxford University Press}
\appnotes{}

\usepackage{enumerate}
\usepackage{xspace}
\usepackage{xcolor}
\usepackage{url}
\usepackage[colorlinks,citecolor=blue]{hyperref}
\usepackage{amsmath}
\usepackage{amssymb}
\theoremstyle{definition}
\newtheorem{defn}{Definition}[]
\newtheorem{task}{Task}[section]

\usepackage[margin=0.95in]{geometry}

\providecommand{\e}[1]{\ensuremath{\times 10^{#1}}}
\DeclareMathOperator*{\argmax}{arg\,max}

\newcommand{\medusa}{{\sc{Medusa}}\xspace}
\makeatother

\begin{document}

\firstpage{1}

\subtitle{Subject Section}

\title[Jumping across biomedical contexts using compressive data fusion]{Jumping across biomedical contexts using compressive data fusion}
\author[Zitnik and Zupan]{Marinka Zitnik\,$^{\text{\sfb 1,2,}*}$ and Blaz Zupan\,$^{\text{\sfb 2,3,}*}$}
\address{
$^{\text{\sf 1}}$Department of Computer Science, Stanford University, CA, USA, and 
$^{\text{\sf 2}}$Faculty of Computer and Information Science, University of Ljubljana, Ljubljana, Slovenia, and 
$^{\text{\sf 3}}$Department of Molecular and Human Genetics, Baylor College of Medicine, TX, USA}

\corresp{$^\ast$To whom correspondence should be addressed.}

\history{}

\editor{}

\abstract{
\textbf{Motivation:} 
The rapid growth of diverse biological data allows us to consider interactions between a variety of objects, such as genes, chemicals, molecular signatures, diseases, pathways and environmental exposures. Often, any pair of objects---such as a gene and a disease---can be related in different ways, for example, directly via gene-disease associations or indirectly via functional annotations, chemicals and pathways. Different ways of relating these objects carry {\em different semantic meanings}. However, traditional methods disregard these semantics and thus cannot fully exploit their value in data modeling.
\\[1mm]
\textbf{Results:} 
We present \medusa, an approach to detect size-$k$ modules of objects that, taken together, appear most significant to another set of objects. \medusa operates on large-scale collections of heterogeneous data sets and explicitly distinguishes between diverse data semantics. It advances research along two dimensions: it builds on collective matrix factorization to derive different semantics, and it formulates the growing of the modules as a submodular optimization program. \medusa is flexible in choosing or combining semantic meanings and provides theoretical guarantees about detection quality. In a systematic study on 310 complex diseases, we show the effectiveness of \medusa in associating genes with diseases and detecting disease modules. We demonstrate that in predicting gene-disease associations \medusa compares favorably to methods that ignore diverse semantic meanings. We find that the utility of different semantics depends on disease categories and that, overall, \medusa recovers disease modules more accurately when combining different semantics. 
\\[1mm]
\textbf{Availability and implementation:} Source code is at \href{http://github.com/marinkaz/medusa}{http://github.com/marinkaz/medusa}.
\\[1mm]
\textbf{Contact:} \href{marinka@cs.stanford.edu}{marinka@cs.stanford.edu}, \href{blaz.zupan@fri.uni-lj.si}{blaz.zupan@fri.uni-lj.si}
}

\maketitle

\section{Introduction}

\begin{figure}
\centerline{\includegraphics[width=0.5\textwidth]{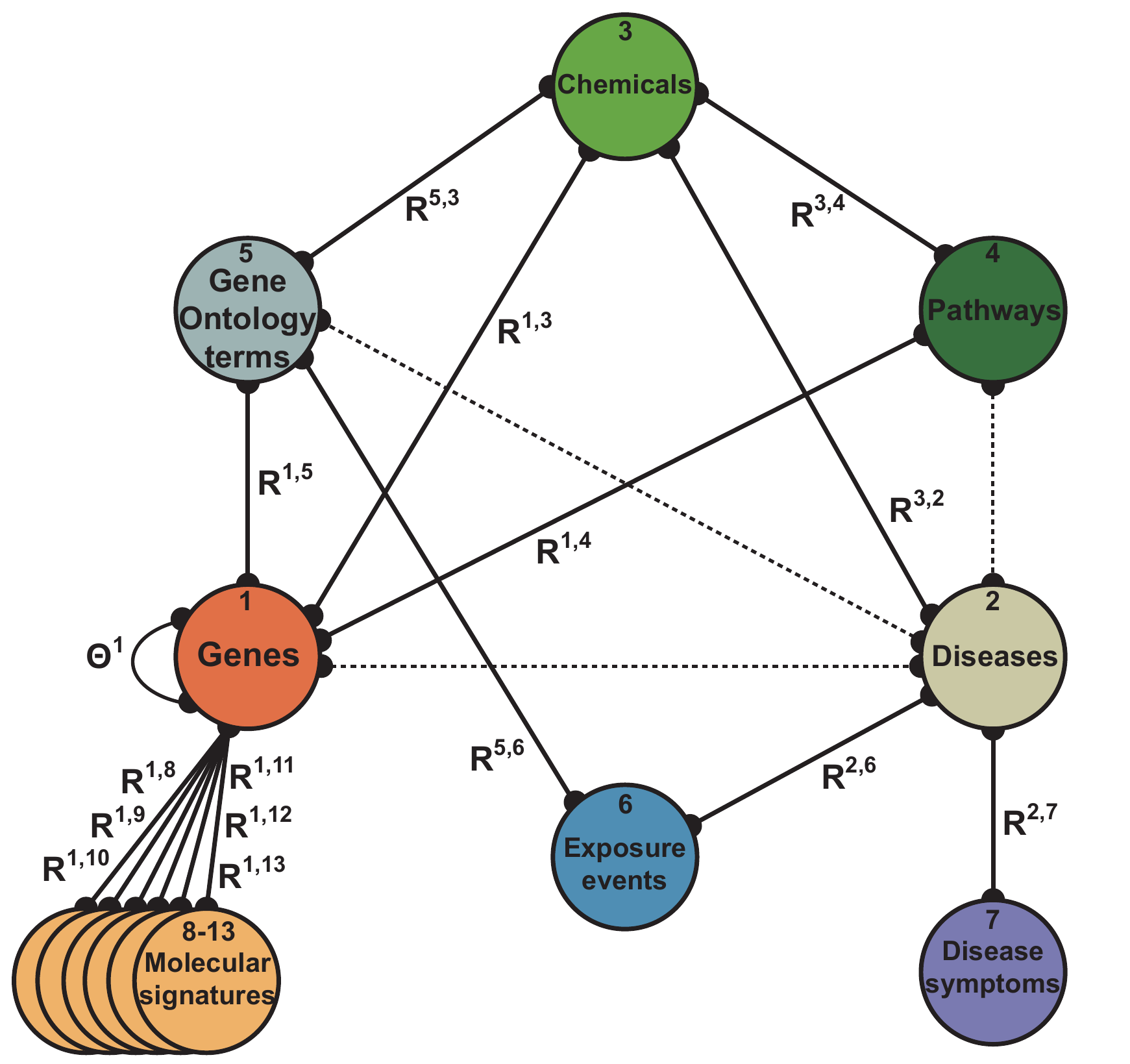}}
\caption{Data fusion graph showing relations between the data sets used in this study. Each node represents a distinct type of objects, such as chemicals, pathways or exposure events, and each edge represents a data set. For example, $\mathbf{R}^{2,6}$ is a matrix of curated exposure data containing environment-disease connections from the CTD database~\citep{Davis2015}. In total, the analysis based on this graph considers objects of 13 different types and 16 data sets (ignoring dotted edges, see Sec.~\ref{sec:data-gene-disease-associations}). For a detailed description of the data, see Sec.~\ref{sec:data}.}\label{fig:fusion-graph}
\end{figure}

In recent years, there is increasing evidence that gene-disease association prediction and disease module detection can benefit from integrative data analysis~\citep{Moreau2012,Ritchie2015}. Large-scale molecular biology data systems analyzed with integrative approaches are typically heterogeneous and contain objects of different types, such as genes, pathways, chemicals, disease symptoms and exposure measurements. These objects interconnect through multiple, most often pairwise, relations encoded in the data. Consider an example of such a data system from Fig.~\ref{fig:fusion-graph} that contains 16 data sets (solid edges) and objects of 13 different types (nodes). For example, data set $\mathbf{R}^{2,7}$ encodes the clinical manifestations of diseases, whereas data set $\mathbf{R}^{5,3}$ describes associations of chemicals with biological processes and molecular functions. The ubiquity of complex data systems of this kind presents many unique opportunities and challenges for uncovering genotype-phenotype interactions, or, in general, interactions between any kind of objects.


Challenges in the joint consideration of systems of data sets, such as that in Fig.~\ref{fig:fusion-graph}, include inferring accurate models to predict disease traits and outcomes, elucidating important disease genes and generating insight into the genetic underpinnings of complex diseases~\citep{Barabasi2011,Han2013,Tacsan2015,Ruffalo2015}. We would like these models to collectively consider the breadth of available data, from whole-genome sequencing to transcriptomic, methylomic and metabolic data~\citep{Navlakha2010,Greene2015,Zitnik2015ploscb}. A major barrier preventing existing methods from fully exploiting entire data collections is that individual data sets usually cannot be directly related to each other. For example, data sets $\mathbf{R}^{5,3}$ (annotation of chemicals with GO terms) and $\mathbf{R}^{2,7}$ (symptoms of diseases) in Fig.~\ref{fig:fusion-graph} reside in completely different feature space. As we will learn in this manuscript, we can relate distant objects by chaining through the fusion graph; for example, we can relate molecular signatures with disease symptoms through genes, GO terms, chemicals and diseases. But such chaining actually exacerbates the problem, as distant objects can be linked in many different ways. For example, another way to relate molecular signatures with disease symptoms is via genes, pathways, chemicals, GO terms~\citep{Ashburner2000}, exposure events and diseases. {\em A priori}, it is not obvious which of the two ways, or which of any existing ways of connecting the signatures with the symptoms, performs better and should thus be preferred when mining disease data.


Different ways of relating objects often carry different {\em semantic meanings} and can potentially generate different results. Intuitively, different semantics imply different similarities. However, traditional methods which we review in the next section disregard the subtlety of different types of objects and links. These methods mix, discard or ignore different semantics, which might impede their performance and explanatory capabilities. In this work, we aim to fill this gap by developing an approach for disease module detection that can consider diverse semantics in a principled manner.


We here introduce a novel approach, called \medusa, for automatic detection of size-$k$ significant modules from heterogeneous systems of biological data. Unlike previous works in integrative data analysis, \medusa explicitly takes different semantics into consideration during module detection by allowing a user to either choose a particular semantic or combine them. Our goal is to answer association queries on possibly complex data systems, such as the one in Fig.~\ref{fig:fusion-graph}. For example, given a small number of diseases, infer the most significant group (module) of genes of size $k$. Or, given a list of genes, propose a group of $k$ other genes that, taken together, will give the highest significance under a particular null hypothesis. Or, given a selection of molecular pathways of interest, find which $k$ chemicals have the largest collective impact on these pathways. To achieve this level of versatility, \medusa builds upon a recent collective matrix factorization algorithm~\citep{Zitnik2015tpami}. In addition, \medusa formulates a submodular optimization program, which provides theoretical guarantees about the significance of the detected modules~\citep{Fujishige2005}. 

In a case study with data sets shown in Fig.~\ref{fig:fusion-graph}, we applied \medusa to find gene-disease associations and infer disease modules. We demonstrate that \medusa-inferred associations are more accurate than those of alternative approaches, which conflate distinct semantics that exist in the data system. Importantly, we find that different semantics vary in their ability to make accurate predictions. We also show that the performance of different semantics depends on the disease category. Finally, we observe that the overall best performance is achieved when \medusa infers associations by combining distinct semantics.


\section{Related work}\label{sec:related-work}

The question of distinguishing different semantics that exist within biomedical data systems remains largely unexplored. Two notable exceptions include a {\em meta-path-based} approach for gene-disease link prediction in heterogeneous networks~\citep{Himmelstein2015} and a {\em latent-chain-based} approach for gene prioritization~\citep{Zitnik2015ploscb}. These approaches, however, are algorithmically different. The approach of \cite{Himmelstein2015} is a network-based technique that relies on {\em meta-paths}~\citep{Sun2011vldb,Sun2011asonam,Sun2012,Wan2015}. Meta-paths represent the number of path instances between two objects that follow a particular sequence of object types in a heterogeneous network. In contrast to meta-paths, \cite{Zitnik2015ploscb} use collective matrix factorization~\citep{Zitnik2015tpami} to estimate a latent data representation of a data system and then derive new connections by appropriately multiplying the latent matrices. In Sec.~\ref{sec:experimental-comparison} we empirically compare our \medusa, which formulates module detection on top of latent representation of the system, to alternative meta-path-based approaches. 

Advances in computational approaches for mining disease related relationships, such as {\em gene-disease, drug-disease or disease-disease associations} may lead to better understanding of human disease and may help identify new disease genes, drug targets and biomarkers~\cite{Barabasi2011}. Representative studies include \cite{Kohler2008,Warde2010,Li2010,Davis2011,Goncalves2012,Zitnik2013scirep}. \cite{Navlakha2010,Barabasi2011} found that random walk approaches usually outperform clustering and neighborhood approaches when predicting gene-disease associations from network data, although most methods make unique predictions not proposed by any other method. Recently, latent factor models (e.g., \cite{Zitnik2013scirep,Natarajan2014,Zitnik2016psb}) have been successful in predicting gene-disease and disease-disease associations. These methods can combine heterogeneous data for diseases and genes by estimating latent models that are coupled across different data sets and explain well the observed associations. Cofunction~\citep{Ruffalo2015,Tacsan2015} networks are also important for fine-scale mapping of diseases by prioritizing genes located at disease-associated loci, for example, by connectivity to known causal genes. Most of these approaches are restricted to inferring pairwise associations. 

There are several lines of research on how to consider many data sets to derive good {\em groupwise disease associations.} \cite{Vanunu2010,Ghiassian2015} considered network propagation and random walk analysis for prioritizing disease genes and inferring protein complex associations. \cite{Han2013} sought groupwise disease associations for sets of single nucleotide polymorphisms mapping to a given functional category. On a related note, guilty-by-association methods~\citep{Wang2012} have used cofunction networks to assign functions to uncharacterized genes in various organisms and to functionally characterize whole sets of genes. Functionally coherent subnetworks were also used to augment curated functional annotations by connecting genes that share, or are likely to share, functions~\citep{Lee2004,Greene2015,Tacsan2015}, for example, by sharing protein domain annotations or tissue-specific interactions. 

However, while these gene-disease association and disease module detection methods use the information from different data sources, none, including our previous work on this topic~\citep{Zitnik2013scirep,Zitnik2016psb}, explicitly considers that different purposes of disease-related analysis might benefit from considering different, potentially distant, semantics between genes and diseases, nor do they ask users to select/combine different ways of connecting genes with diseases. \medusa, the algorithm presented here, differs from the above methods in that it utilizes semantically distinct chains consisting of possibly many relations to derive relationships between objects, such as genes and diseases. \medusa can establish connections between objects for which direct relationships are not available in present data. It then uses these connections to find size-$k$ modules of objects that together exhibit near-highest significance to the preselected pivot objects. The flexibility of \medusa is further shown in that the object type of the pivots can, but not necessarily, coincide with the object type of the candidates and that chains carrying different semantic meanings can be combined in a principled manner.

\begin{methods}

\section{Methods}

\medusa is an approach for the detection of size-$k$ modules that are maximally significant for a predefined set of pivot objects. On the input, \medusa accepts (1) a set of candidate objects that are potential module members, (2) a set of pivot objects that are not necessarily of the same type as the candidates and (3) a possibly large and heterogeneous collection of data sets represented in the form of a data fusion graph such as that shown in Fig.~\ref{fig:fusion-graph}.

\medusa uses collective matrix factorization to jointly estimate a latent data model from all data sets included in a fusion graph. It exploits the latent data model to \emph{establish semantically distinct connections} between candidate objects and domains of other object types in the fusion graph (Sec.~\ref{sec:terminology-setup}). The module detection algorithm in \medusa (Sec.~\ref{sec:submodularity-in-medusa}--\ref{sec:medusa-combining}) is a {\em submodular optimization} program which yields an efficient algorithm and provides theoretical guarantees about the significance of the detected modules. 

\subsection{Preliminaries and Notation}

We start by briefly introducing a data fusion graph, which is a way to represent a collection of data sets to be examined in an integrative way. We then overview collective matrix factorization that estimates a latent data representation of the fusion graph in \medusa. Finally, we review important mathematical concepts from submodular optimization, which are needed in the derivation of the \medusa module detection algorithm. 

\subsubsection{Data fusion graph}

A data fusion graph $\mathcal{G} = (\mathcal{V}, \mathcal{R}, \mathcal{T})$ is a relational map of data sets~\citep{Zitnik2015tpami}. Nodes of the graph $\mathcal{V}$ represent different types of objects, such as ontological terms, genes, diseases, pathways and chemicals. The edges of the graph correspond to data sets, which are given in matrices annotated next to the edges. An exemplar data fusion graph is shown in Fig.~\ref{fig:fusion-graph}. Matrix $\mathbf{R}^{2,7}$ therein is a $n_2 \times n_7$ real-valued matrix whose rows correspond to diseases indexed by a respective disease-based controlled vocabulary and whose columns indicate disease symptoms indexed by a symptom-based vocabulary. Elements of matrix $\mathbf{R}^{2,7}$ represent a data set, such as disease-symptom associations.

Technically, the edges of the fusion graph are given by a set of {\em relation matrices} $\mathcal{R} = \{\mathbf{R}^{I, J}; I, J \in \mathcal{V}, I \neq J, \mathbf{R}^{I, J} \in \mathbb{R}^{n_I \times n_J}\}$ that represent dyadic data sets and a set of {\em constraint matrices} $\mathcal{T} = \{\mathbf{\Theta}^{I}; I \in \mathcal{V}, \mathbf{\Theta}^{I} \in \mathbb{R}^{n_I \times n_I}\}$ that represent unary data sets. It is possible to have multiple relation matrices that relate object types $I$ and $J$ (i.e., more than one edge between $I$ and $J$ in the fusion graph) or multiple constraint matrices for object type $I$ (i.e., more than one loop for $I$ in the fusion graph). Here, this possibility is suppressed for notational brevity. 


\subsubsection{Collective matrix factorization}

Collective matrix factorization~\citep{Zitnik2015tpami} is an algorithm that considers a fusion graph $\mathcal{G} = (\mathcal{V}, \mathcal{R}, \mathcal{T})$ and infers its latent model by compressing the data sets with co-factorization of matrices in $\mathcal{R}.$ Matrices $\mathcal{T}$ are used for regularization of the latent model. The method simultaneously co-factorizes all the relation matrices into the products of much smaller latent matrices through a procedure which (i) ensures the {\em transfer of information between related matrices} and (ii) promotes good generalization via {\em a high-quality data compression}. To achieve the first point, collective factorization reuses the latent matrices when decomposing distinct but related relation matrices. The second feature is possible due to the low-dimensional nature of matrix factorization. 

The collective matrix factorization algorithm aims to estimate low-dimensional latent matrices $\mathbf{G}^I$, $I \in \mathcal{V}$, and $\mathbf{S}^{I,J}$, $I, J \in \mathcal{V}$, which minimize the following objective:
\begin{equation}
\sum_{\mathbf{R}^{I,J} \in \mathcal{R}} \!\!\!\!\!\| \mathbf{R}^{I,J} - \mathbf{G}^I \mathbf{S}^{I,J} (\mathbf{G}^J)^T \|_{\text{Fro}}^2 + \sum_{\mathbf{\Theta}^I \in \mathcal{T}} \!\! \text{tr}((\mathbf{G}^I)^T \mathbf{\Theta}^I \mathbf{G}^I).
\label{eq:dfmf-objective}
\end{equation}
Here, the inferred latent matrices tri-factorize each relation matrix as $\widehat{\mathbf{R}}^{I,J} = \mathbf{G}^I \mathbf{S}^{I,J} (\mathbf{G}^I)^T.$ Matrix $\mathbf{G}^i$ is a $n_I \times k_I$ ($k_I \ll n_I$) non-negative latent matrix containing latent profiles of objects of type $I$ in rows, $G^J$ is a $n_J \times k_J$ ($k_J \ll n_J$) non-negative latent matrix with profiles of objects of type $J$ in rows and $\mathbf{S}^{I,J}$ is a $k_I \times k_J$ latent matrix that models interactions between latent components in the $(I, J)$-th data set. The latent profile of an object of type $I$ is given by its corresponding row vector in $\mathbf{G}^I.$ Semantically, the profile encodes membership of the object to the $k_I$ latent components.

The parameters of the algorithm are factorization ranks, $k_I$, for every object type $I$ in the data fusion system, which are selected as in \cite{Zitnik2015ploscb}. We refer the reader to \cite{Zitnik2015tpami} for a detailed description and theoretical analysis of the factorization algorithm.

\subsubsection{Chaining of latent data matrices}

A factorized system of latent data matrices returned by the collective matrix factorization can be used to establish connections between distant object types, i.e., non-neighboring nodes in the fusion graph~\citep{Zitnik2015ploscb}.

\begin{defn}
{\em \textbf{Chain.} A chain $\mathcal{C}^{S,T}$ is a sequence of relations defined on a fusion graph $\mathcal{G} = (\mathcal{V},\mathcal{R}, \mathcal{T})$ that connects object type $S \in \mathcal{V}$ with possibly distant object types $T \in \mathcal{V}.$ The chain $\mathcal{C}^{S,T}$ is denoted in the form of:
\begin{equation}
\mathcal{C}^{S,T} = \mathbf{R}^{S,I_1} \circ \mathbf{R}^{I_1, I_2} \circ \dots \circ \mathbf{R}^{I_{l-2}, I_{l-1}} \circ \mathbf{R}^{I_{l-1}, T},
\label{eq:chain-path}
\end{equation}
which defines a composite relation between object types $S$ and $T$, where $\circ$ denotes the composition operator on relations. Here, for $I_1, I_2, \dots, I_l$ the object types $I_j$ and $I_{j+1}$ must be adjacent in the fusion graph $\mathcal{G}$.}
\end{defn}

The length $l$ of chain $\mathcal{C}^{S,T}$ is measured by the number of its constituent relations. We now formulate how to materialize a given chain and derive the profiles of objects of one type in the space of objects of another type.

\begin{defn}
{\em \textbf{Materialized chain.}
Given a fusion graph $\mathcal{G} = (\mathcal{V}, \mathcal{R}, \mathcal{T})$ and a latent data system estimated by collective matrix factorization, a materialized chain $\mathbf{C}^{S,T} \in \mathbb{R}^{n_S \times n_T}$ for chain $\mathcal{C}^{S,T}$ specified in Eq.~\eqref{eq:chain-path} is defined as:
\begin{equation}
\mathbf{C}^{S,T} = \widehat{\mathbf{R}}^{S,I_1} \widehat{\mathbf{R}}^{I_1, I_2}  \dots \widehat{\mathbf{R}}^{I_{l-2}, I_{l-1}} \widehat{\mathbf{R}}^{I_{l-1}, T},
\label{eq:chain-materialize}
\end{equation}
where $\widehat{\mathbf{R}}^{I,J}$ is the relation matrix reconstructed from the latent data system as $\widehat{\mathbf{R}}^{I,J} = \mathbf{G}^I \mathbf{S}^{I,J} (\mathbf{G}^{J})^T.$ ($(\cdot)^T$ is matrix transposition.)}
\label{def:materialized}
\end{defn}

\subsubsection{Submodular functions and optimization}\label{sec:submodular-preliminaries}


\begin{figure}
\centerline{\includegraphics[width=0.5\textwidth]{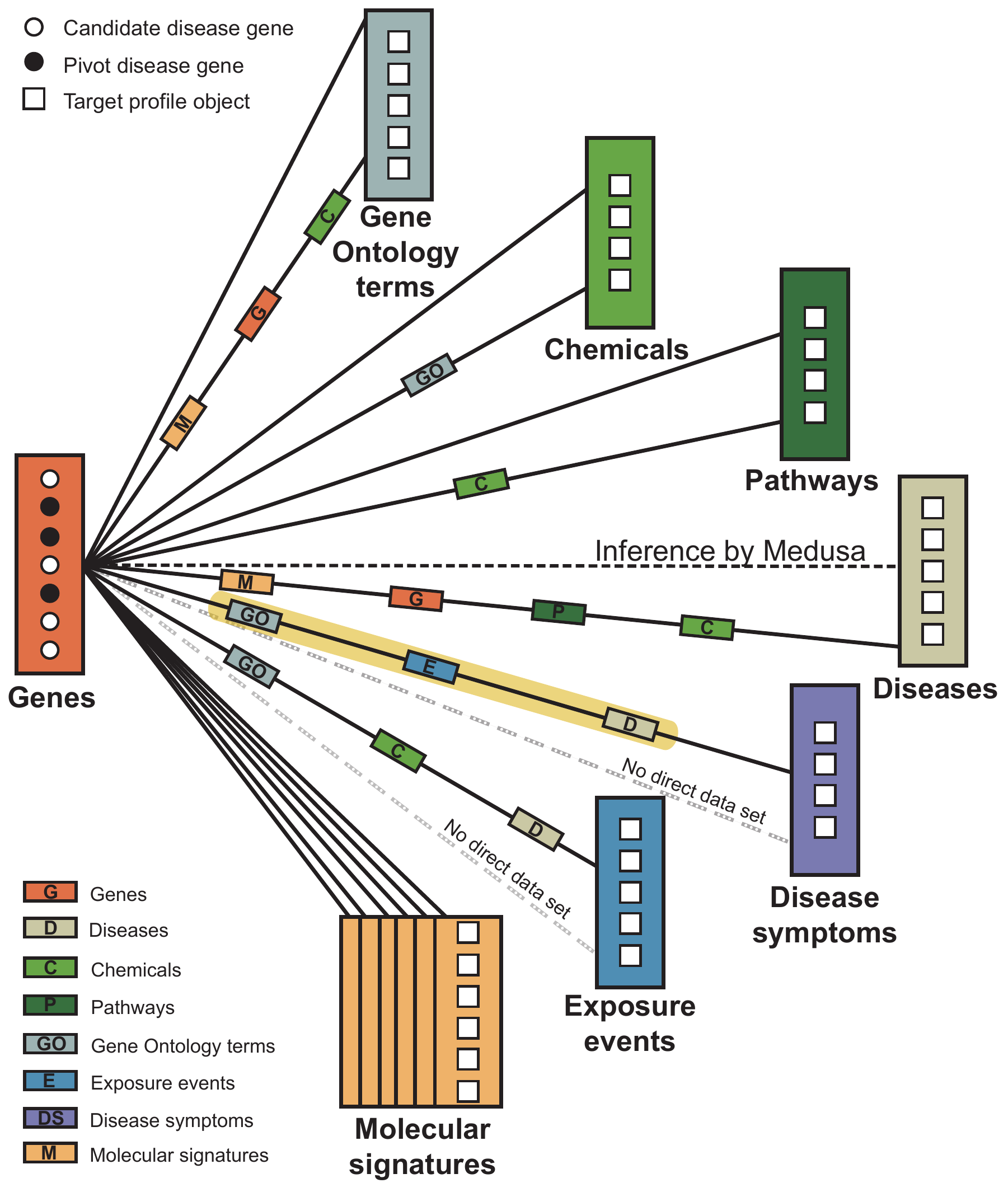}}
\caption{The CPE regime in \medusa. In the CPE environment, \medusa detects a relevant module of candidate objects based on a set of pivot objects, which belong to the same object type as the candidates. For example, given three disease genes (pivots, black circles), we want to find other potentially relevant disease genes (candidates, white circles), a task denoted with a black dashed line. In a special case where the studied objects are genes, as shown here, and we are interested in size-$1$ modules, the CPE regime coincides with the well known gene prioritization task. The figure shows 15 distinct semantic aspects (solid black lines) that exist in the fusion graph in Fig.~\ref{fig:fusion-graph} to relate genes with all other types of objects. For example, one semantic to relate genes with disease symptoms goes through Gene Ontology terms (``GO''), exposure events (``E'') and diseases (``D''). Notice we cannot directly relate genes to disease symptoms, i.e., at least one other object type is needed to establish the connection (grey dashed line). }\label{fig:cp-type-equivalence}
\end{figure}

Submodular functions~\citep{Edmonds1970,Fujishige2005} have recently attracted much interest, e.g., see \cite{Krause2011}. Let us assume we are given a finite set of $n$ objects $V$ and a valuation function $f: 2^{V} \rightarrow \mathbb{R}_+$ that returns a non-negative real value for any subset $X\subseteq V.$ The function $f$ is said to be {\em submodular} if it satisfies the property of diminishing returns. That is, for any set $X \subseteq Y$ and $i \not\in Y$, we must have:
$f(X \cup \{i\}) - f(X) \ge f(Y \cup \{i\}) - f(Y).$
This means that the incremental gain of element $i$ decreases when the background in which $i$ is considered grows from $X$ to $Y \supseteq X.$ We define the ``gain'' as $f(i, X) = f(X \cup \{i\}) - f(X)$, which implies that $f$ is submodular if $f(i, X) \ge f(i, Y).$

In this paper we deal with functions that are not only submodular but also non-negative (i.e. $f(X) \ge 0$ for all $X \subseteq V$) and monotone non-decreasing (i.e. $f(X) \ge f(Y)$ for all $X \subseteq Y$). Such functions are trivial to uselessly maximize, since $f(V)$ is the largest possible valuation. However, we would typically like to identify a valuable subset of bounded and small cost. Here, we are interested in subsets whose costs are measured by their size. This leads to the optimization problem $X^* \in \argmax_{X \subseteq V, |X| \le k} f(X),$ where $k$ is the desired subset size. Solving this problem exactly is NP-complete~\citep{Feige1998}. However, when $f$ is submodular, then the greedy algorithm has a worst case guarantee of $f(\widetilde{X}^*) \ge (1 - 1/e) f(X_{\text{opt}}) \approx 0.63 f(X_{\text{opt}})$, where $X_{\text{opt}}$ is the optimal and $\widetilde{X}^*$ is the greedy solution~\citep{Nemhauser1978}.

\subsection{Problem definition}\label{sec:terminology-setup}

In this section, we introduce a framework for module detection on data fusion graphs, a novel approach to find size-$k$ maximally significant modules, \medusa, and propose a \medusa-based top-$k$ module detection problem that takes into consideration diverse semantics in heterogeneous data systems. 

We start by defining the concepts needed to guide the module detection procedure and to assess the significance of the modules. 
\begin{defn}
{\em \textbf{Candidate objects.} Given a fusion graph $\mathcal{G} = (\mathcal{V}, \mathcal{R}, \mathcal{T})$, candidate objects are given by a set $I = \{i_1, i_2, \dots, i_{n_I}\}$ of all the entities that belong to type $I \in \mathcal{V}.$ Candidates constitute a pool of objects from which a module is identified.}
\end{defn}
\begin{defn}
{\em \textbf{Pivot objects.} Given a fusion graph $\mathcal{G} = (\mathcal{V}, \mathcal{R}, \mathcal{T})$, pivot objects are given by a subset $S_0 = \{s_1, s_2, \dots, s_u \},$ $u < n_J$, of the entities of type $J \in \mathcal{V}.$ Pivots are the objects against which the significance of the current \medusa module is assessed.}
\end{defn}
Depending on whether the pivots and the candidates belong to the same or different type of objects in $\mathcal{G}$, we distinguish two prediction settings. This distinction is important because it will lead to different optimization objectives when detecting modules in Sec.~\ref{sec:detection}.
\begin{task} \label{task:medusa}
We aim to find a size-$k$ module $M_k = \{m_1, m_2, \dots, m_k \}$ of the candidates $I$ that display the maximal significance with respect to the given pivots $S_0$.\\
Let $J$ be the object type of the pivots. 
(1) In the \textbf{CPE} (candidate-pivot-equivalence) \textbf{regime}, candidates and pivots are of the same data type, $I = J.$ (2) In the \textbf{CPI} (candidate-pivot-inequivalence) \textbf{regime}, candidates and pivots are of different types, $I \neq J.$
\end{task}

We illustrate the CPE and the CPI regimes with concrete examples in Fig.~\ref{fig:cp-type-equivalence} and Fig.~\ref{fig:cp-type-inequivalence}, respectively. We proceed by formulating a measure, which we optimize for when detecting \medusa modules.

\begin{figure}
\centerline{\includegraphics[width=0.5\textwidth]{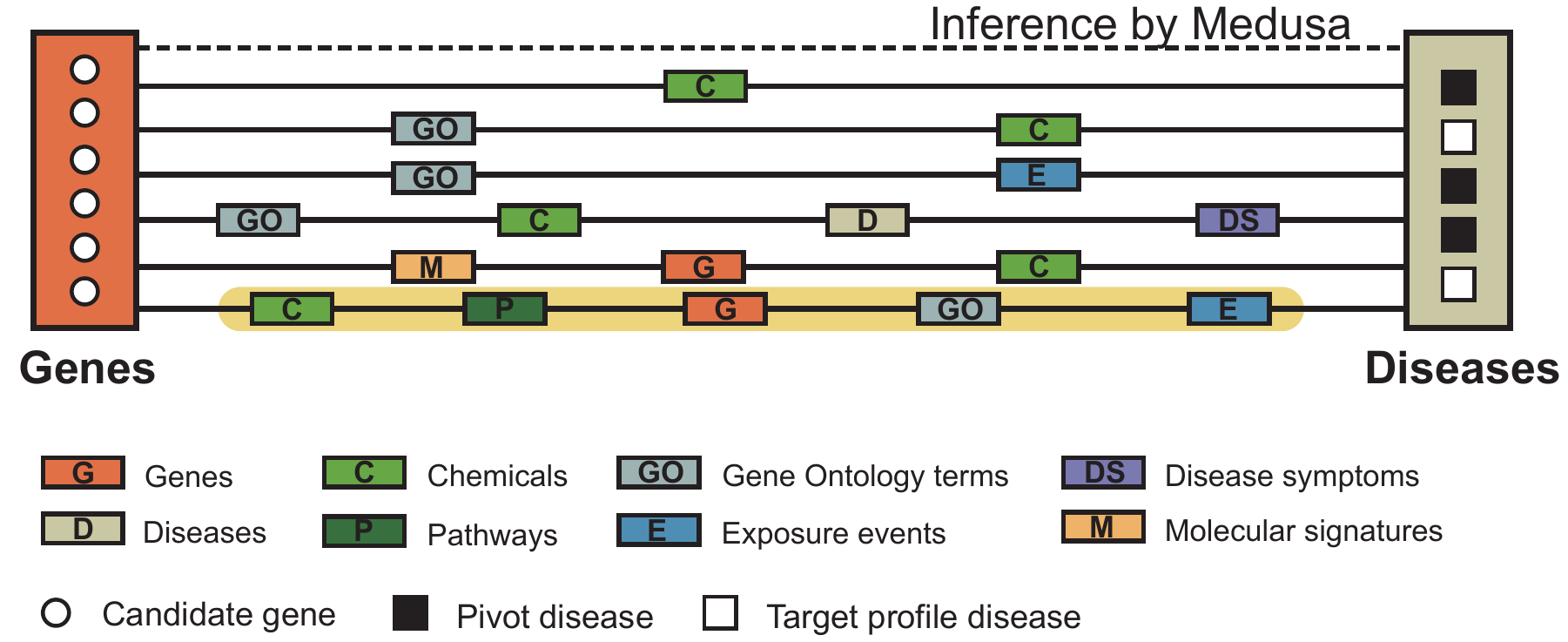}}
\caption{The CPI regime in \medusa. In the CPI regime, \medusa detects a relevant module of candidate objects based on a set of pivot objects, which belong to a different object type than the candidates. For example, given three diseases (pivots, black squares), we want to find potentially relevant genes (candidates, white circles), a task denoted with a dashed line. The figure shows 6 distinct semantics (solid lines) that exist in the fusion graph in Fig.~\ref{fig:fusion-graph}, which \medusa can choose or combine to identify the relevant module. For example, highlighted is an aspect that connects genes with diseases via chemicals (``C''), pathways (``P''), genes (``G''), Gene Ontology terms (``GO'') and exposure events (``E''). }\label{fig:cp-type-inequivalence}
\end{figure} 

\subsection{Submodularity for detection of Medusa modules}\label{sec:submodularity-in-medusa}

Submodularity is a natural model for detection of size-$k$ maximally significant modules in multiplex data. In this case, each $i \in V_{\text{source}}$ is a distinct candidate and $V_{\text{source}}$ corresponds to a set of all candidate objects. An important characteristic of a good model for this problem is that we wish to decrease the ``value'' of a candidate $i \in V_{\text{source}}$ based on how much that candidate has in common with candidates $S_r$ that have been chosen in the first $r$ rounds. 

The value $p(i, S_r)$ of a given candidate $i$ in a background of previously chosen objects $S_r \subseteq V_{\text{source}}$ further diminishes as the background grows $V_{\text{source}} \supseteq S_t$, $t > r.$ When, for example, both candidate and pivot objects are genes and a candidate's value is represented as the statistical significance of its concentration, it is natural for the significance to be discounted based on how much representation of that candidate already exists in a previously chosen subset. When the module grows it naturally becomes more diverse and hence its characterization is less distinctive, which results in the overall reduction of statistical power. This means that the candidate is pulled into the module when its significance towards the pivot objects is the highest. If the candidate were added to the module later, its significance could only be smaller. That is, if we were to observe that candidate after being included into the module, its significance would fade into insignificance. This paradigm corresponds to submodularity, which we express mathematically by functions in Eq.~(\ref{eq:signif-cpe}) and Eq.~(\ref{eq:signif-cpi}) below. 

\subsection{Detection of size-$k$ maximally significant modules}\label{sec:detection}

Next, we describe the \medusa module detection algorithm. Recall that \medusa is able to operate in two prediction regimes defined in Sec.~\ref{sec:terminology-setup}. We start by describing the algorithm for the CPE regime and proceed with the algorithm for the CPI regime. 

Recall that a particular semantic aspect connecting object types $S$ and $T$ is realized as matrix $\mathbf{C}^{S,T}$ (Definition~\ref{def:materialized}). For notational convenience we here denote a given aspect simply as a matrix $\mathbf{C}.$ Prior to the analysis, matrix $\mathbf{C}$ is normalized according to row sums in $\mathbf{C},$ which yields a row-stochastic matrix. 

\subsubsection{Medusa in the CPE regime}\label{sec:cpe-medusa}

We capture the distinct connectivity patterns of the candidates by evaluating the {\em significance} of their connections in matrix $\mathbf{C}.$ For a randomly picked candidate we evaluate the probability that a certain fraction of the candidate's strongest connections to the objects in the columns of matrix $\mathbf{C}$ match exactly with the strongest connections of the pivots. 

In the simplest case, we would simply count the connections, and this notion would correspond to the hypergeometric distribution. However, since matrix $\mathbf{C}$ is a real-valued object inferred by a latent model, we take into account the estimated strength of connectivity rather than its mere existence (cf. experiments in Sec.~\ref{sec:experimental-comparison}). For this, we need to extend the binomial coefficient to the real line using the gamma function~\citep{Fowler1996}. Technically, this is implemented in the following definition. 

\begin{defn}
{\em \textbf{Candidate concentration.}
Given a semantic $\mathbf{C},$ pivots $S_0$, and a candidate $i$, we define the concentration of candidate $i$ as:
\begin{align}
h_{{\rm CPE}}(c, S_0, Q_i) & = {\rm Bin}\Big(\!\! {\displaystyle \sum_{q \in Q_i}} \mathbf{C}_{S_0 q}^T \mathbf{w}_{S_0}, c \Big) \times \nonumber \\
& \!\!\!\!\!\!\!\!\!\!\!\!\!\! {\rm Bin}\Big({\displaystyle \sum_{j}} \mathbf{C}_{S_0 j}^T \delta(\mathbf{w}_{S_0}) - \!\!\!\!\! {\displaystyle \sum_{q \in Q_i}} \!\! \mathbf{C}_{S_0 q}^T \mathbf{w}_{S_0}, \mathbf{C}_{i} \mathbf{1} - c \Big) \times \nonumber \\
& \!\!\!\!\!\!\!\!\!\!\!\!\!\! 1/{\rm Bin}\Big({\displaystyle \sum_{j}} \mathbf{C}_{S_0 j}^T \delta(\mathbf{w}_{S_0}), \mathbf{C}_{i}\mathbf{1}\Big), 
\label{eq:opt-cpe}
\end{align}
where $c$ is the observed strength of connectivity of candidate $i$, $\text{Bin}(n, k) = \Gamma(n+1)/(\Gamma(k+1) \Gamma(n-k+1)),$ and $\Gamma(x)$ is the gamma function evaluated at $x.$ Here, $\delta(x)$ is the indicator function, $\delta(x) = 1$ if $x = 1$ and $\delta(x) = 0$ otherwise. Also, $\mathbf{X}_{S_0}$ returns the rows of $\mathbf{X}$ indexed by set $S_0.$ Here, $Q_i$ is a set containing column indices of $i$'s strongest connections. The weights in vector $\mathbf{w}$ is defined as $\mathbf{w}_i = 1$ if $i \in S_0$, else $\mathbf{w}_i = (1 - \alpha)^r$ if $i \in S_r ,$ where $r$ is the iteration when $i$ was included into the module, and $\mathbf{w}_i = 0$ otherwise. The value for $\alpha,$ which promotes modules that are tight around the initial set of pivots $S_0$, and the size of $Q_i$, are user-defined parameters. 
}
\end{defn}

We evaluate whether candidate $i$ has greater correspondence with the pivots than expected under this null hypothesis by calculating the cumulative probability for the observed or any weaker concentration of the connectivity:
\begin{equation}
p_{{\rm CPE}}(i, S_0; Q_i) \triangleq {\displaystyle \int_{|Q_i|}^{ \widetilde{c}_i }} h_{{\rm CPE}}(c, S_0, Q_i) dc
\label{eq:signif-cpe}
\end{equation}
where $\widetilde{c}_i = \mathbf{C}_{i Q_i} \mathbf{1}.$ A better candidate will have a lower $p_{{\rm CPE}}$ value.

It can be shown that the use of this cumulative probability to select a candidate object, which is to be included into the current module, leads to a submodular program (the proof is omitted here for brevity). This appealing property allows us to use the greedy algorithm to find modules of size $k$ that approximate the optimal modules within a constant factor (Sec.~\ref{sec:submodular-preliminaries}). Recall that the optimal modules can only be found using a prohibitively expensive exhaustive enumeration of all size-$k$ modules. Building on these observations, we propose the algorithm to identify a maximally significant module of size $k$, which solves Task~\ref{task:medusa} in the CPE regime:
\begin{enumerate}[i. ]
\item Start with an empty module $M_0 = \{ \}$. 
\item Compute concentration significance (Eq.~\eqref{eq:signif-cpe}) for all candidates.
\item Rank the candidates according to their respective $p_{\rm CPE}$ values.
\item Add the candidate $i$ with the highest rank (i.e. lowest $p_{\rm CPE}$ value) to the current \medusa module, $M_{r} \rightarrow M_{r+1} = M_{r} \cup \{i\}$, and to the set of pivot objects, $S_{r} \rightarrow S_{r+1} = S_{r} \cup \{ i\}.$
\item Repeat steps ii.--iv. $k$-times with the expanded set of pivot objects.
\end{enumerate}
The order in which the candidates are pulled into the module reflects their relevance to the pivot objects according to semantic $\mathbf{C}.$ Fig.~\ref{fig:medusa-equivalence} is an example for finding a size-3 gene module based on a semantic that relates genes to disease symptoms via Gene Ontology terms, exposure events and diseases. 


\subsubsection{Medusa in the CPI regime}\label{sec:cpi-medusa}

So far, we have seen that in the CPE regime, candidates and pivots are given by the rows of matrix $\mathbf{C}$. However, in the CPI regime this is not true anymore. Here, pivots correspond to columns of matrix $\mathbf{C},$ whereas candidates are still given by rows of matrix $\mathbf{C}.$ To adjust for this change, we assess a candidate's significance by its visibility, which we define next. 

\begin{figure}
\centerline{\includegraphics[width=0.5\textwidth]{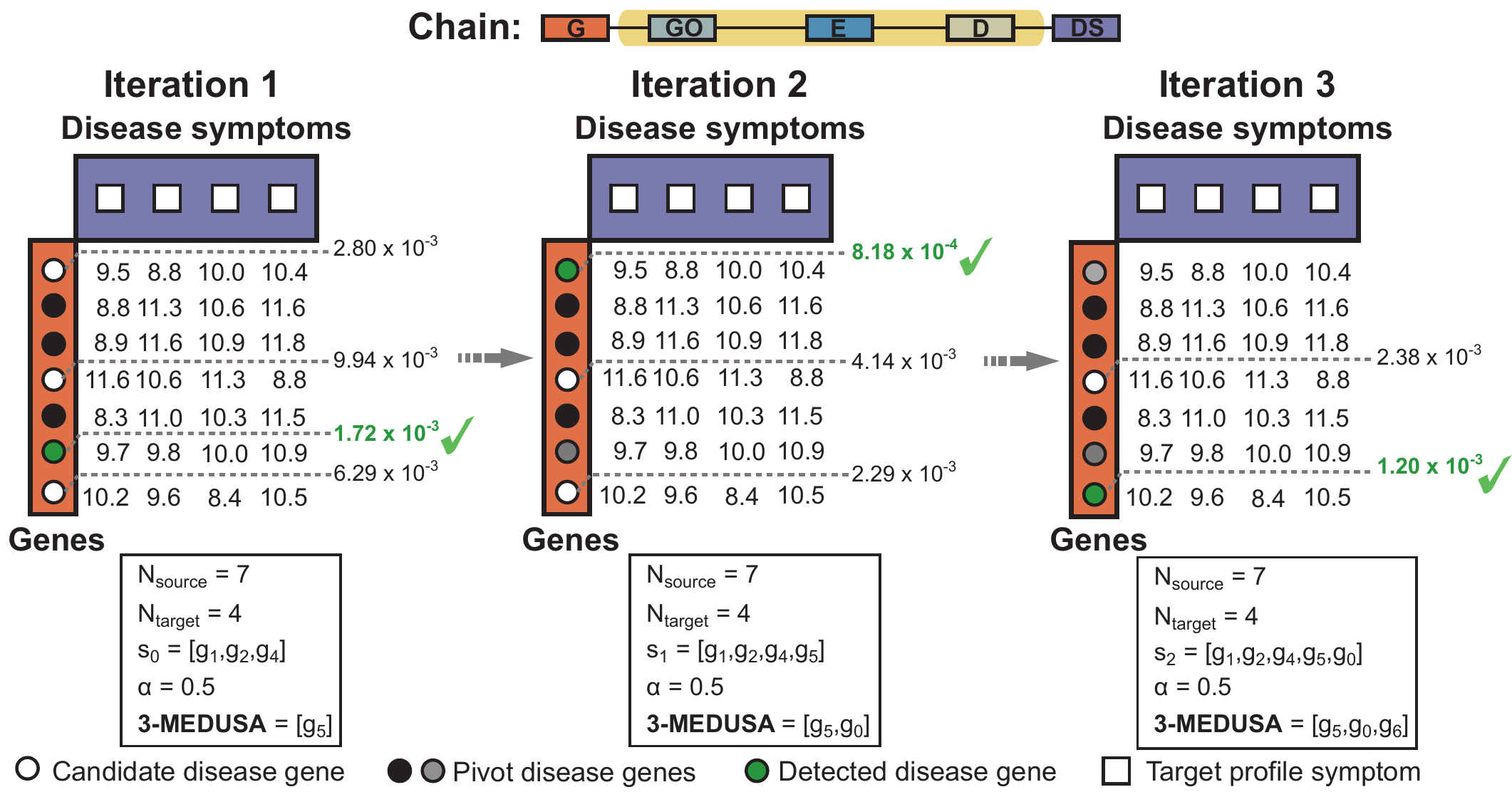}}
\caption{Inferring a $3$-maximally significant distant module with \medusa in the CPE regime. Shown is a toy example of a chain that relates 7 genes in rows to 4 disease symptoms in columns via Gene Ontology terms, exposure data and diseases (see the highlighted chain in Fig.~\ref{fig:cp-type-equivalence}). Given the three pivot genes shown as black circles, we would like to identify the most significant size-3 gene module. Notice that we operate in the CPE regime where both pivot and candidate objects are of the same type (i.e. genes). In the first iteration, candidate $g_5$ achieves the lowest score (i.e. $p_{\text{CPE}} = 1.72\e{-3}$) and is thus added to the module and included into the pivot set. However, the importance of $g_5$ as a pivot object is downweighted according to the $\alpha = 0.5$. Notice also that candidate $g_3$ has a predominantly reversed concentration relative to the pivot genes $g_1, g_2, g_4$ in the first iteration, which results in the poor score of $g_3$ (i.e. $p_{\text{CPE}} = 9.94\e{-3}$). In particular, candidate $g_3$ is concentrated on symptoms $d_0$ and $d_2$, whereas pivot genes are concentrated on $d_1$ and $d_3$. The score for $g_3$ improves in later iterations when the \medusa module becomes more diverse. The final module is $M_3 = \{g_5, g_0, g_6\}.$}\label{fig:medusa-equivalence}
\end{figure}

\begin{defn}
{\em \textbf{Candidate visibility.} 
Given a semantic $\mathbf{C},$ pivots $S_0$ and a candidate $i$, we define the visibility of candidate $i$ as:

\begin{align}
h_{{\rm CPI}}(c, S_0) & = {\rm Bin}\Big( \sum_{j} \mathbf{C}_{j S_0} \mathbf{1}, c \Big) \times \nonumber \\
& {\rm Bin}\Big(  \sum_{l, j} \mathbf{C}_{lj} - \sum_{j} \mathbf{C}_{j S_0} \mathbf{1}, \mathbf{C}_i \mathbf{1} - c \Big) \times \nonumber \\
& 1/{\rm Bin}\Big( \sum_{l, j} \mathbf{C}_{lj}, \mathbf{C}_i \mathbf{1} \Big).
\label{eq:opt-cpi}
\end{align}
The notation follows that in Eq.~(\ref{eq:opt-cpe}).
}
\end{defn}

Intuitively, the visibility of a candidate is the strength of its connections with the pivots. We evaluate whether candidate $i$ has stronger connections to the pivots than expected under this null hypothesis by calculating the cumulative probability for observed or any stronger visibility:
\begin{equation}
p_{{\rm CPI}}(i, S_0) \triangleq {\displaystyle \int_{ \widetilde{c}_i }^{m}} h_{{\rm CPI}}(c, S_0) dc
\label{eq:signif-cpi}
\end{equation}
where $\widetilde{c}_i = \mathbf{C}_{i Q_i} \mathbf{1}$ and $m$ is the number of columns in $\mathbf{C}.$ A better candidate will have a lower $p_{{\rm CPI}}$ value.

Similarly as for the CPE regime, the use of cumulative probability $p_{{\rm CPI}}$ leads to a {\em submodular optimization program}, which has the same appealing properties as the cumulative probability $p_{{\rm CPE}}$ in Sec.~\ref{sec:cpe-medusa}. Building on the theory of submodular functions, we tackle Task~\ref{task:medusa} in the CPI regime by proposing the following greedy algorithm to identify a maximally significant module of size $k$:
\begin{enumerate}[i. ]
\item Start with an empty module $M_0 = \{ \}$.
\item Compute visibility significance (Eq.~\eqref{eq:signif-cpi}) for all candidates. Visibility of candidate $i$ naturally decreases in iteration $r$ according to
$\widetilde{c}_i = (1 - \sum_{t=1}^r \beta^t \exp(-\text{KL}(\mathbf{C}_i, \mathbf{C}_{x_{t-1}}))) \mathbf{C}_{i S_0} \mathbf{1},$
where KL denotes the Kullback-Leibler divergence, $\mathbf{C}_{x_{t-1}}$ is matrix row of the candidate selected in iteration $t-1$ and $\beta$ is a user-defined parameter promoting diverse modules.
\item Rank the candidates according to their respective $p_{\rm CPI}$ values.
\item Add the candidate $i$ with the highest rank (i.e. lowest $p_{\rm CPI}$ value) to the current \medusa module, $M_{r} \rightarrow M_{r+1} = M_{r} \cup \{i\}$.
\item Repeat steps ii.--iv. $k$-times, bringing in one candidate object at a time into the growing module. 
\end{enumerate}
Fig.~\ref{fig:medusa-inequivalence} shows a toy example for finding a size-3 gene module based on a semantic that relates genes with diseases via chemicals, pathways, genes, Gene Ontology terms and exposure events. 

%
%

\begin{figure}
\centerline{\includegraphics[width=0.5\textwidth]{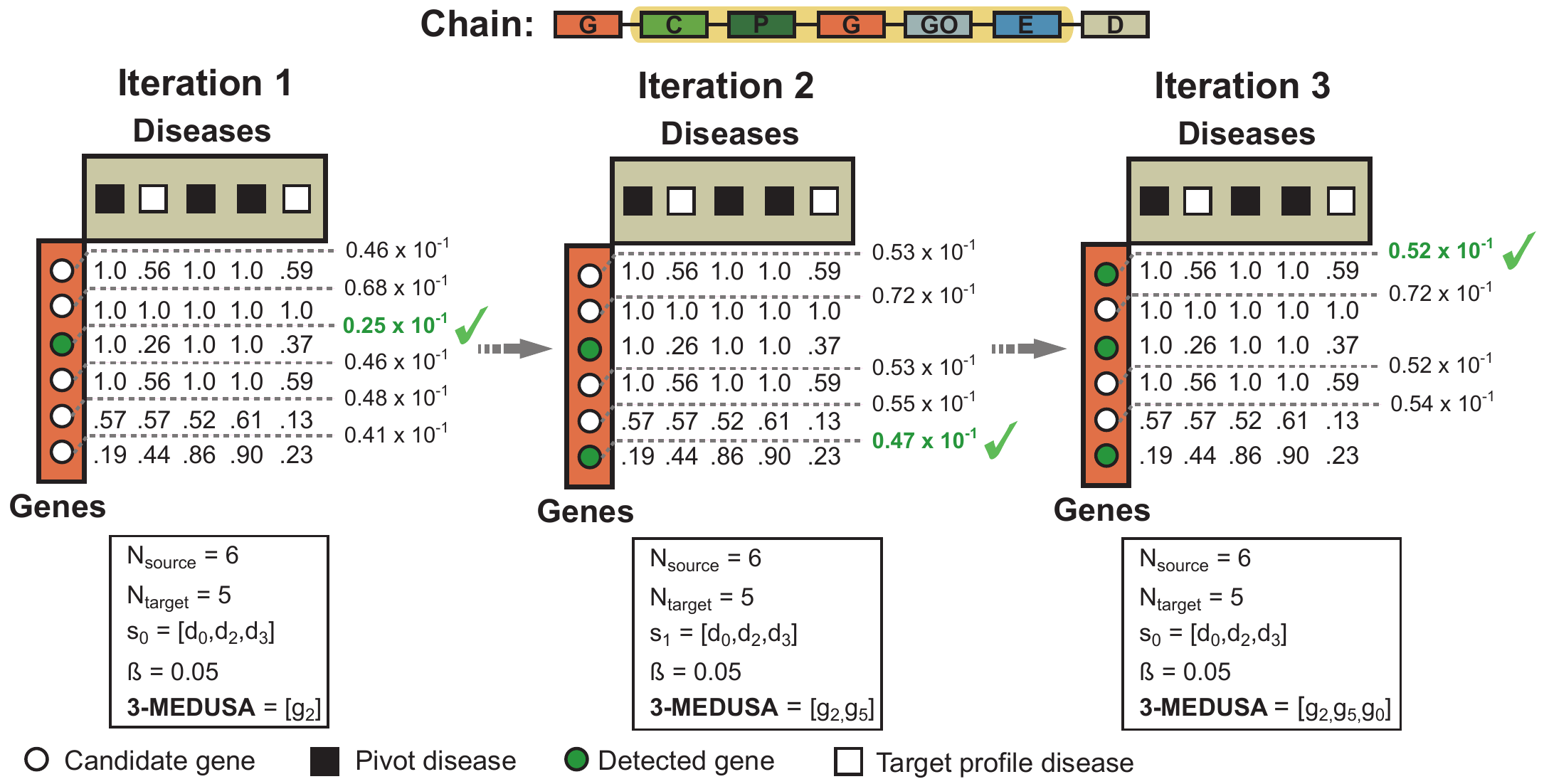}}
\caption{Inferring a $3$-maximally significant distant module with \medusa in the CPI regime. Shown is an example of a chain that relates 6 genes to 5 diseases via chemicals, pathways, genes, Gene Ontology terms and exposure data (see the highlighted chain in Fig.~\ref{fig:cp-type-inequivalence}). In this toy example we are given three pivot diseases shown in black squares and would like to find a $3$-maximally significant gene module. Notice that we operate in the CPI data regime because the pivots (i.e. diseases) are of different type than the candidates (i.e. genes). In the first iteration, gene $g_2$ shows the most significant visibility for the pivot diseases (i.e. $p_{\text{CPI}} = 0.25\e{-1}$) and is thus included into the module. Gene $g_1$ does not discriminate between the pivot and non-pivot diseases and is hence considered an unlikely candidate (i.e. $p_{\text{CPI}} = 0.68\e{-1}$ in the first iteration and $p_{\text{CPI}} = 0.72\e{-1}$ in later iterations). In second iteration, the algorithm picks $g_5$, although one might expect that $g_1$ would be selected due to its distinctive connections to the pivot disease. This is because \medusa detects modules that are not only highly visible to the pivot objects but are also diverse, which is important when trying to identify non-redundant comprehensive modules. Such behavior of \medusa is regulated by parameter $\beta$, which promotes diverse modules in this example, $\beta = 0.05$. The final module is $M_3 = \{g_2, g_5, g_0\}.$
}\label{fig:medusa-inequivalence}
\end{figure}

\subsection{Combining chains carrying different semantics}\label{sec:medusa-combining}

So far, we have described the \medusa algorithm that operates on one particular semantic given by a single matrix $\mathbf{C}$. To be able to combine $d$ different semantics, given by a set of chained matrices $\mathbf{C}^{(1)}$, $\mathbf{C}^{(2)}, \dots, \mathbf{C}^{(d)}$, we employ the following technique. 

In the $r$-th iteration of \medusa, we independently score the candidates according to Eq.~(\ref{eq:signif-cpe}) (in the case of the CPE regime) or Eq.~(\ref{eq:signif-cpi}) (in the case of the CPI regime) for all matrices $\mathbf{C}^{(1)}$, $\mathbf{C}^{(2)}, \dots, \mathbf{C}^{(d)}.$ We then combine candidate scores from different semantics into one score per candidate by an affine combination of semantics' weights. 

In the CPE regime, this is done such that the semantics which rank the pivots higher are assigned larger weights than those in which the pivots are ranked lower. Intuitively, this means that semantics that are more informative for the given set of pivots contribute more towards the final candidate score. 

In the CPI regime, the combination of semantics is done such that the semantics in which the pivots have more similar profiles measured by  the KL divergence are assigned larger weights than those in which the pivots have more heterogeneous data profiles. Detailed steps are provided within the online implementation of \medusa. 

Once the aggregated candidate scores are calculated, \medusa uses algorithms from Sec.~\ref{sec:cpe-medusa} and Sec.~\ref{sec:cpi-medusa} to detect the modules.

\end{methods}

\section{Experimental setup}

We next describe sixteen data sets that were considered in our experiments. We also introduce the ground-truth information against which we evaluated the performance and conclude with a description of the experimental procedure, parameter selection and measures of accuracy. 

\subsection{Data sets in the data fusion graph}\label{sec:data}

In our experiments we consider a collection of data sets shown in Fig.~\ref{fig:fusion-graph}. Table~\ref{tab:data-sets} lists public sources from which data were obtained to build 16 data matrices. 

Gene Ontology (GO)~\citep{Ashburner2000} annotations were downloaded from http://geneontology.org in December 2015 containing 481,685 human gene product annotations. UniProt protein accession identifiers were collapsed to human NCBI's Entrez gene identifiers using the mapping provided by the HGNC resource~\citep{Gray2015}. Curated human protein-protein interactions were retrieved from the BioGRID 3.4.131 database~\citep{Chatr2014}. The recourse contained interactions for 19,702 genes. Data on clinical manifestation of diseases were obtained from the Human symptoms-disease network (HSDN)~\citep{Zhou2014} (Suppl. data S4) and included 147,978 relationships between symptoms and diseases. Term co-occurrences between symptoms (MeSH Symptom terms) and diseases (MeSH Disease terms) were weighted by the TF-IDF values. 

We also compiled gene sets from the Molecular Signatures Database (MSigDB)~\citep{Subramanian2005} in December 2015: 326 positional gene sets (MSig-C1) corresponding to each human chromosome and each cytogenetic bands; 3,395 curated gene sets (MSig-C2) representing expression signatures of genetic and molecular perturbations; 1,330 gene sets (MSig-C2) corresponding to canonical representations of biological processes from the pathway databases; 186 gene sets (MSig-C2) derived from the KEGG database; 221 motif gene sets (MSig-C3) with genes sharing microRNA binding motifs; and 615 gene sets (MSig-C3) with genes sharing transcription factor binding sites. Curated chemical-gene interactions ($\mathbf{R}^{1,3}$), chemical-pathway interactions ($\mathbf{R}^{3,4}$), gene-pathway associations ($\mathbf{R}^{1,4}$), chemical-function associations ($\mathbf{R}^{5,3}$), chemical-disease ($\mathbf{R}^{3,2}$), function-exposure events ($\mathbf{R}^{5,6}$), exposure event-disease relationships ($\mathbf{R}^{2,6}$) and gene-disease relationships were retrieved from the Comparative Toxicogenomics Database (CTD)~\citep{Davis2015} in December 2015.

Each data set was represented with a real-valued data matrix as indicated in Table~\ref{tab:data-sets}. Prior to the analysis, all matrices  were independently column-row normalized according to the second vector norm. 

\subsection{Disease modules and gene-disease associations}\label{sec:data-gene-disease-associations}

The corpus of 310 diseases was downloaded from the CTD~\citep{Davis2015} with the criteria that every disease should have at least 10 and at most 100 curated gene associations for which direct evidence is available in the CTD database. These diseases and their associated genes constituted our ground-truth information against which we evaluated disease module detection and gene-disease association prediction. On average, each disease had 28 associated genes. 

To ensure there was no leakage of information from training to test set in our integrative analysis, we used the following protocol. From our data system we excluded data sets that directly or indirectly rely on disease-gene associations (dotted lines in Fig.~\ref{fig:fusion-graph}). For example, we skipped the pathway-to-disease data set (a hypothetical data matrix $\mathbf{R}^{4,2}$ in Fig.~\ref{fig:fusion-graph}) because a particular pathway would only be linked to a particular disease if there was a disease-associated gene in this pathway. Another example of an excluded data set is the Gene Ontology term-to-disease data set (a hypothetical data matrix $\mathbf{R}^{5,2}$ in Fig.~\ref{fig:fusion-graph}). Here, a given term and a given disease would be linked only if there existed a gene that was both annotated with the term in the Gene Ontology and associated with the disease. 

\begin{table}[t]
\setlength{\tabcolsep}{2pt}
\processtable{Data sets used for the analyses presented in this paper. The ``Reference'' columns list last names of first authors of relevant publications. \label{tab:data-sets}} {
{\scriptsize \begin{tabular}{lcc|lcc}\toprule 
Matrix & Reference & Size & Matrix & Reference & Size \\\midrule
$\mathbf{R}^{1,5}$ & Ashburner & $19{,}828 \times 19{,}951$ & $\mathbf{R}^{5,6}$ & Davis & $19{,}951 \times 565$\\
$\mathbf{R}^{1,3}$ & Davis & $19{,}828 \times 12{,}614$ & $\mathbf{R}^{1,8}$ & Subramanian & $19{,}828 \times 326$\\
$\mathbf{R}^{1,4}$ & Davis & $19{,}828 \times 291$ & $\mathbf{R}^{1,9}$ & Subramanian & $19{,}828 \times 3{,}395$\\
$\mathbf{R}^{2,6}$ & Davis & $9{,}350 \times 565$ & $\mathbf{R}^{1,10}$ & Subramanian & $19{,}828 \times 186$\\
$\mathbf{R}^{2,7}$ & Zhou & $9{,}350 \times 321$ & $\mathbf{R}^{1,11}$ & Subramanian & $19{,}828 \times 1{,}330$\\
$\mathbf{R}^{3,2}$ & Davis & $12{,}614 \times 9{,}350$ & $\mathbf{R}^{1,12}$ & Subramanian & $19{,}828 \times 221$\\
$\mathbf{R}^{3,4}$ & Davis & $12{,}614 \times 291$ & $\mathbf{R}^{1,13}$ & Subramanian & $19{,}828 \times 615$\\
$\mathbf{R}^{5,3}$ & Davis & $19{,}951 \times 12{,}614$ & $\mathbf{\Theta}^{1}$ & Chatr-aryamontri & $19{,}828 \times 19{,}828$\\\botrule
\end{tabular}}}{}
\end{table}

This protocol enabled us to construct a data system for performance evaluation of various methods that was not contaminated by existing gene-disease associations. For example, data set $\mathbf{R}^{3,2}$ in Fig.~\ref{fig:fusion-graph} contained curated chemical-disease associations that were extracted from the published literature by the CTD curators~\citep{Davis2015}. However, this data set omitted inferred chemical-disease associations from our training data, which associated chemicals and diseases through shared gene interactions. 

\subsection{Performance evaluation}

We evaluated prediction accuracy using a disease-centric cross-validation procedure. (1) Prediction of gene-disease associations was evaluated as follows. For a particular disease in our corpus of 310 diseases, we conducted a leave-one-gene-out cross-validation to obtain an estimated score for the left-out gene. The remaining (training) genes were considered positive instances and were used to select and fit the model parameters. (2) Detection of disease modules was evaluated using the procedure from \cite{Ghiassian2015}. For each disease module, we randomly removed a certain fraction (25\%, 50\% and 75\%) of the disease genes and used the remaining genes as pivots.

For \medusa, we need to specify the parameters required by collective matrix factorization, the value for $\alpha$, the size of $Q$ (in the case of the CPE regime) and the value for $\beta$ (in the case of the CPI regime). We tuned $\alpha$, $\beta$ and the size of $Q$ in an internal cross-validation procedure on the training genes. Factorization ranks for collective matrix factorization were selected using a procedure similar to the one described in \cite{Zitnik2015ploscb}; 13 values were required, each representing latent dimension of one object type in our fusion graph. We selected these dimensions through a single parameter $p$, which specified latent dimension for an object type as a fraction of the number of objects of that type: $k_I = p n_I.$ The value $p=0.05$ was used in the experiments because it maximized the mean AUPRC achieved on a set of 10 diseases from the CTD, which were later not considered for performance comparison. The selection of parameters for other approaches was made based on internal cross-validation. 

We measured accuracy using the areas under precision-recall curves (AUPRC) and under receiver operating characteristic curves (AUROC).

\section{Results and discussion}

\subsection{Capturing different biological semantics with Medusa}

\begin{figure*}
\centerline{\includegraphics[width=\textwidth]{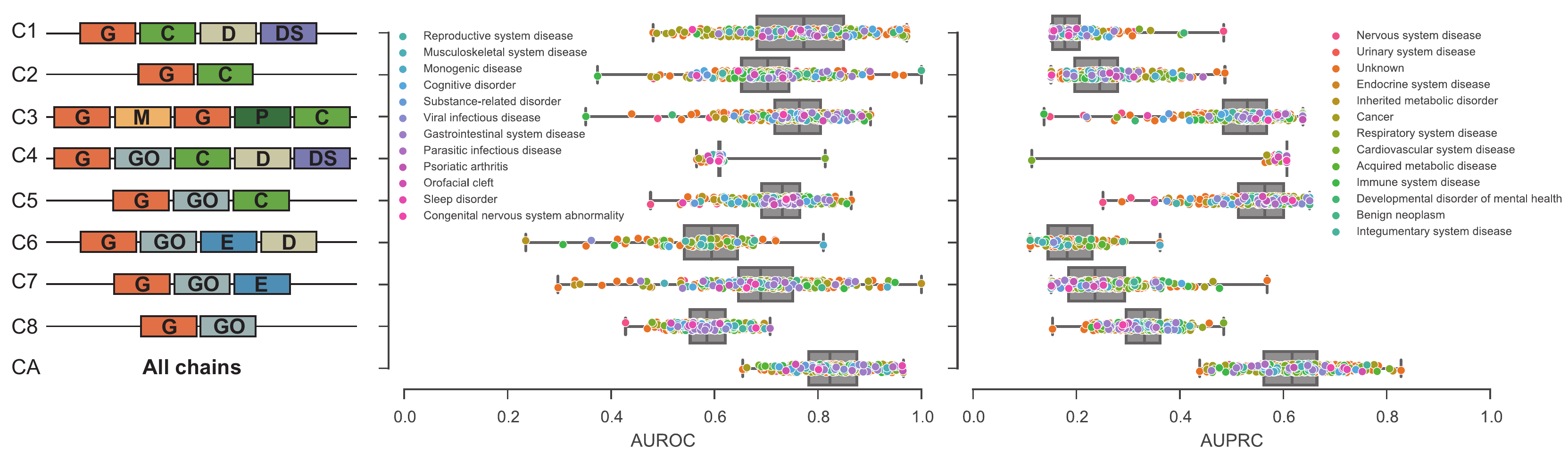}}
\caption{Gene-disease association prediction with \medusa. Nine different biological semantics (C1--C8, CA) were considered in the analysis. Each semantic is shown as a sequence of object types contained in the fusion graph (Fig.~\ref{fig:fusion-graph}). For example, in the ``C4'' semantic, \medusa estimated gene-disease associations based on the latent chain that related genes (``G'') with disease symptoms (``DS'') via Gene Ontology terms (``GO''), chemicals (``C''), diseases (``D'') and disease symptoms (``DS''). For each distinct biological semantic we report the AUROC and AUPRC values aggregated over 310 complex diseases. For visualization purposes, diseases were partitioned into 25 disease classes based on Disease Ontology~\citep{Kibbe2014}, such as ``reproductive system diseases'' and ``musculoskeletal system diseases,'' and the points representing accuracy scores of each individual disease are colored according to corresponding disease classes. Notice the substantial variation of performance across different semantics. Generally, \medusa achieved the highest accuracy when combining semantics from C1--C8 (i.e., CA, last row). }\label{fig:res-associations}
\end{figure*}

First, our goal was to investigate the effects of different biological semantic aspects on the prediction of gene-disease associations. We used \medusa to estimate disease genes for 310 diseases included in the CTD database (Sec.~\ref{sec:data}). We considered eight distinct semantic aspects denoted as C1--C8 (Fig.~\ref{fig:res-associations}, left). For example, chain C6 corresponds to a matrix that relates genes to diseases via Gene Ontology terms and exposure events (see Definition~\ref{def:materialized}). 

\medusa detects significant modules of a specified size $k$. To apply it to the prediction of gene-disease associations, we search for size-1 modules and use probability estimates returned by the algorithm (Eq.~(\ref{eq:signif-cpe})) to make  predictions.

Fig.~\ref{fig:res-associations} shows the performance of \medusa in terms of AUROC and AUPRC. In addition to the eight distinct semantics, we also analyzed the performance of \medusa in its mode which combines different semantic aspects, denoted as CA in the figure (Sec.~\ref{sec:medusa-combining}). We further studied how prediction accuracy varies across classes of diseases, which were identified using Disease Ontology~\citep{Kibbe2014}. A cut off at level 2 of the Disease Ontology graph revealed 25 disease classes (Fig.~\ref{fig:res-associations}), such as ``cognitive disorders'' and ``gastrointestinal diseases.''  

We observed substantial variation of AUROC and AUPRC values across different semantic aspects (i.e., chains). In terms of the AUROC, the best single semantic appeared to be C1, which was followed closely by C3 and C5. In addition to genes, chemicals were the common object type considered for construction of these three chains. These results are important because they demonstrate that different ways of establishing connections between genes and diseases can result in more or less accurate predictions. The results also suggest that objects of different types and links carry different semantic meanings, and it might not make sense to mix them without distinguishing their semantics when associating genes with diseases. This experiment also alludes to the {\em explanatory} value of \medusa. \medusa is able to provide insights into the utility of different semantics, a capability, which most present models for co-factorization of multiple matrices do not have. 

While a user might explicitly specify a semantic that he would like to consider in a concrete application, \medusa can also make predictions that are consistent with multiple chains. In particular, combining semantics C1--C8 in Fig.~\ref{fig:res-associations} yielded the most accurate predictions overall. However, prediction accuracy in Fig.~\ref{fig:res-associations} varies greatly by disease and we explore this issue next.

\subsection{Detecting disease modules with Medusa}

\begin{figure}
\centerline{\includegraphics[width=0.5\textwidth]{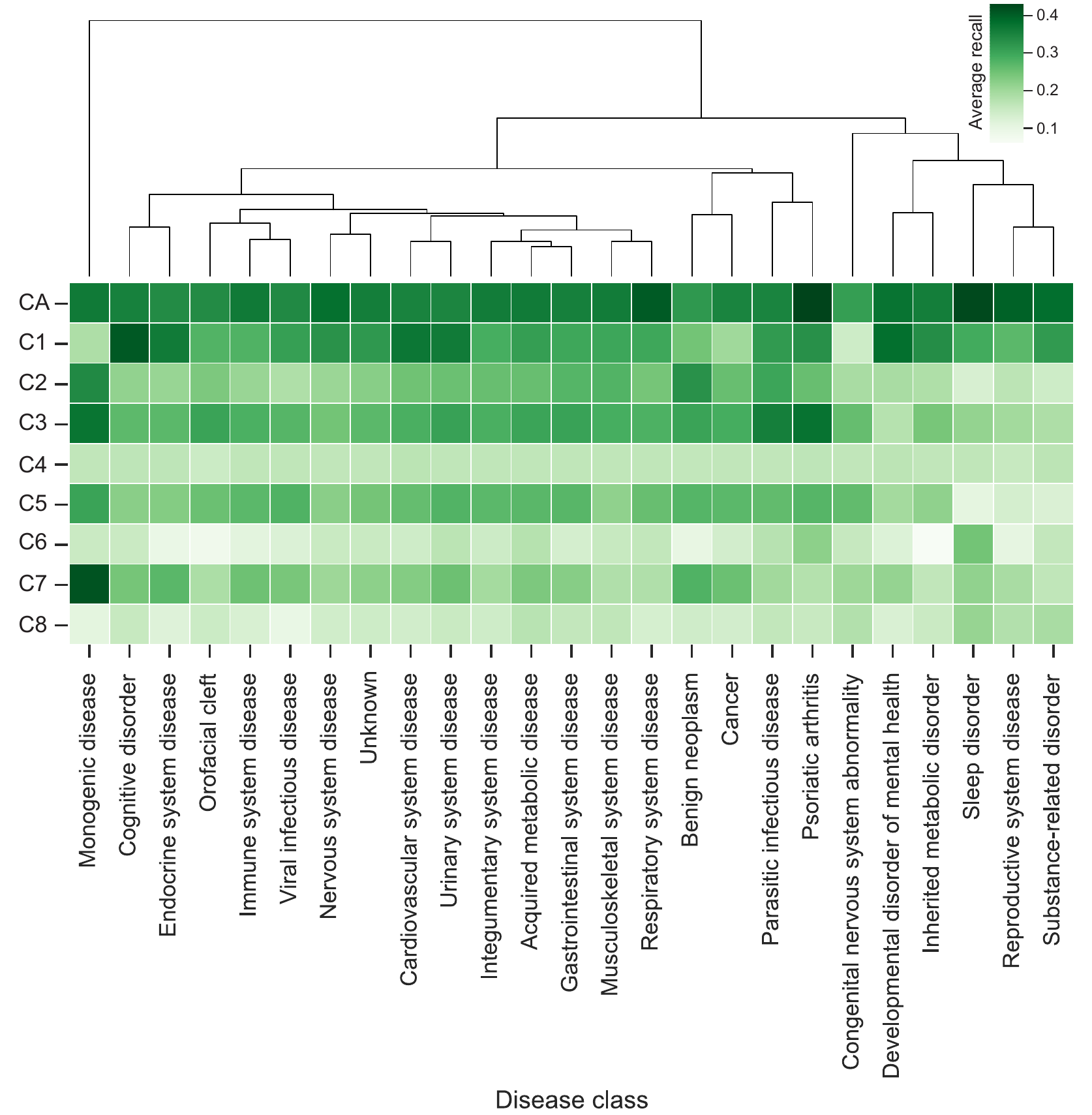}}
\caption{Disease module detection with \medusa. We considered nine different semantic meanings (C1--C8, CA; rows in the heat map) listed in Fig.~\ref{fig:res-associations}. Reported is the recovery rate when 50\% of genes from known disease modules were left out. The recall values were calculated for 310 diseases in our corpus and then average aggregated into 25 groups based on a categorization of diseases in the Disease Ontology~\citep{Kibbe2014} (columns in the heat map). The dendrogram reveals that recovery rates of disease modules from related disease classes are similar.}\label{fig:res-modules}
\end{figure}

We analyzed the extent to which \medusa could recover the full disease module if we removed a certain fraction of disease associated genes. Recall that a disease module is given by the set of genes associated with that disease in the CTD~\citep{Davis2015}. For a given disease, \medusa used 50\% of the disease genes as pivots and detected a size-$k$ module, where $k$ was set to the size of the full disease module. Notice that \medusa pulls in one gene at a time into the growing module and given the inferred latent model, the running time of the module detection increases linearly with the desired module size $k$. Fig.~\ref{fig:res-modules} shows the fraction of held-out disease genes (recall) that were found in \medusa modules. Higher values indicate better performance.

We found that the highest rate of true positives was achieved in the early iterations of the \medusa algorithm, i.e., when the number of executed iterations was less than the size of the full module. This is an important observation because it indicates that the highest ranked genes are most likely to be part of the disease module.

The results in Fig.~\ref{fig:res-modules} further show that the estimated recovery rates varied across different semantic aspects as can be seen by comparing rows in the heat map. Typically, the best performance was observed when \medusa was used to detect modules based on joint analysis of all semantics (see CA in Fig.~\ref{fig:res-modules}).

It is interesting to examine which classes of diseases display higher recovery rates than others and how the rates compare to each other. The dendrogram in Fig.~\ref{fig:res-modules} shows that ``monogenic diseases'' exhibited a distinct recovery pattern. For example, modules corresponding to monogenic disorders according to Disease Ontology~\citep{Kibbe2014} were best recovered using the chain C7, whereas other disease classes (with the exception of cognitive disorders) were best detected when \medusa was used in the CA setting. We also observed that related disease classes displayed similar patterns of recovery rates across different semantics. For example, immune system diseases and viral infectious diseases are placed closely together in the dendrogram, as well as acquired metabolic diseases and diseases of the gastrointestinal system. It is known that diseases from similar disease classes are more likely to be associated with sets of genes that overlap~\citep{Barabasi2011}. The similar recall patterns from related disease classes thus suggest that the \medusa outcome is robust with respect to variations in the set of pivot genes. 


\begin{table}
\setlength{\tabcolsep}{2pt}
\processtable{Cross-validated performance for predicting gene-disease associations using a heterogeneous data system shown in Fig.~\ref{fig:fusion-graph}. Higher values indicate better performance. Reported are averaged values over 310 diseases and the maximum of the upper/lower quartile distances. \label{tab:method-comparison}} {
\begin{tabular}{p{2.7cm}p{2cm}|cc}\toprule 
\multicolumn{2}{c|}{Approach} & AUPRC & AUROC \\
\multicolumn{1}{l}{Data model} & \multicolumn{1}{l|}{Prediction model} & & \\\midrule
Meta-path model & Correlation & $0.339 \pm 0.17$ & $0.599 \pm 0.07$\\
Meta-path model & PathSim & $0.587 \pm 0.18$ & $0.754 \pm 0.13$\\\midrule
Heterogeneous network & Random walk & $0.566 \pm 0.14$ & $0.772 \pm 0.11$\\\midrule
Collective latent model & Correlation & $0.483 \pm 0.23$ & $0.605 \pm 0.09$\\
Collective latent model & Random walk & $0.535 \pm 0.17$ & $0.762 \pm 0.16$\\
\multicolumn{2}{c|}{\medusa$\!\!^*$} & $0.617 \pm 0.21$  & $0.831 \pm 0.14$ \\\botrule
\end{tabular}}{\hspace{-6mm} $^*$The analysis combined eight distinct biological semantics (C1--C8) shown in Fig.~\ref{fig:res-associations}.}
\end{table}

\subsection{Comparing Medusa with existing methods}\label{sec:experimental-comparison}

So far we have studied the utility of \medusa to take into consideration distinct semantics that exist in heterogeneous biological data when predicting gene-disease associations and detecting disease modules. We proceed by examining how \medusa performs relative to several other approaches for mining gene-disease associations. 

First, we compare \medusa with meta-path based approaches (Sec.~\ref{sec:related-work}). These approaches have just recently been tested on prediction problems in biology for the first time~\citep{Himmelstein2015} and have shown promising performance for prioritizing genetic associations from genome-wide association studies. Meta-paths~\citep{Sun2011vldb} are sequences of object types. They are used to represent complex relationships between objects beyond what links in a homogeneous network capture. For example, given a meta-path that corresponds to the chain C6: Genes $\rightarrow$ Gene Ontology terms $\rightarrow$ Exposure events $\rightarrow$ Diseases (Fig.~\ref{fig:res-associations} left), a meta-path-based approach in its simplest form relates a particular candidate gene with a particular pivot disease by counting the number of paths in a heterogeneous network between a candidate and pivot node. These counts then serve to derive features. Each feature represents one meta-path originating in a given candidate gene and terminating in a given pivot disease and quantifies the prevalence of that meta-path between any gene-disease pair. To describe different aspects of connectivity, we computed eight features based on chains C1--C8 (Fig.~\ref{fig:res-associations}, left) and then used a rank-correlation metric or a sophisticated PathSim meta-path-based metric~\citep{Sun2011vldb,Wan2015} to score gene-disease associations. The results in Table~\ref{tab:method-comparison} show that \medusa compares favorably to both meta-path models in terms of AUROC and AUPRC values. It is important to understand the subtlety: \medusa relates candidate genes to diseases by deriving new connections between them based on matrices estimated by a collective latent factor model. This highlights \medusa's advantage of taking into consideration projections of data into the latent space, which potentially give more informative connections than the rather crude meta-path count metrics. Furthermore, \medusa's technique to estimate associations considers the significance of derived connections under a particular null hypothesis, whereas alternative methods rely on similarity scoring. 
 
Second, we applied a random walk algorithm~\citep{Li2010} to the heterogeneous network whose schema is shown in Fig.~\ref{fig:fusion-graph}. It is known that random walk approaches are often the best performing methods for associating genes with diseases~\citep{Navlakha2010}. We found that the random walk approach has performance comparable to the meta-path-based approach that used the PathSim metric (Table~\ref{tab:method-comparison}). However, in the majority of the diseases, \medusa achieved higher cross-validated accuracy. 

Finally, we also considered two simplified variants of \medusa (Table~\ref{tab:method-comparison}, third block). To measure the effect of \medusa's submodular optimization program, we ran \medusa against variations, which associated genes to diseases based on (1) the rank-correlation between candidate gene profile and disease gene profiles in a materialized chained matrix or (2) the gene-disease score returned by the random walk approach. \medusa offered an overall improvement of 37\% over the correlation-based variant and a 10\% improvement over the random walk approach as measured by the AUROC. The results suggest that both key ingredients of \medusa, the collective latent model and the submodular program, are important for its good performance.

\section{Conclusion}

We here presented a novel and practical approach to infer connections between objects that are either close to each other or far away from each other in heterogeneous biological data domains. We introduced \medusa, a module detection algorithm that, given a set of pivot objects, finds a size-$k$ module of candidate objects that are jointly relevant to the pivots. Importantly, this module achieves significance that is provably close to the maximum significance that could be achieved by any size-$k$ set of the candidates. Our experiments reveal the versatility of \medusa to accurately detect disease modules and predict gene-disease associations by either flexibly choosing or combining different semantic meanings. The distinct property of \medusa to distinguish diverse semantics enabled \medusa to compare favorably against several alternative methods. These findings put \medusa on the path towards a biomedical data fusion search engine. 


\section*{Funding}

This work was supported by the ARRS (P2-0209, J2-5480) and the NIH (P01-HD39691). M.Z. was supported in part by the DARPA SIMPLEX program. 

{\noindent\em Conflict of interest:} none declared. 

\bibliographystyle{natbib}
\bibliography{zitnik}

\begin{thebibliography}{}

\bibitem[Ashburner {\em et~al.}(2000)Ashburner {\em et~al.}]{Ashburner2000}
Ashburner, M. {\em et~al.} (2000).
\newblock Gene {O}ntology: tool for the unification of biology.
\newblock {\em Nature Genetics\/}, {\bf 25}(1), 25--29.

\bibitem[Barab{\'a}si {\em et~al.}(2011)Barab{\'a}si, Gulbahce, and
  Loscalzo]{Barabasi2011}
Barab{\'a}si, A.-L., Gulbahce, N., and Loscalzo, J. (2011).
\newblock Network medicine: a network-based approach to human disease.
\newblock {\em Nat. Rev. Genet.}, {\bf 12}(1), 56--68.

\bibitem[Chatr-aryamontri {\em et~al.}(2014)Chatr-aryamontri {\em
  et~al.}]{Chatr2014}
Chatr-aryamontri, A. {\em et~al.} (2014).
\newblock The {BioGRID} interaction database: 2015 update.
\newblock {\em NAR\/}, page gku1204.

\bibitem[Davis {\em et~al.}(2015)Davis {\em et~al.}]{Davis2015}
Davis, A.~P. {\em et~al.} (2015).
\newblock The {C}omparative {T}oxicogenomics {D}atabase's 10th year
  anniversary: update 2015.
\newblock {\em NAR\/}, {\bf 43}(D1), D914--D920.

\bibitem[Davis and Chawla(2011)Davis and Chawla]{Davis2011}
Davis, D.~A. and Chawla, N.~V. (2011).
\newblock Exploring and exploiting disease interactions from multi-relational
  gene and phenotype networks.
\newblock {\em PLoS One\/}, {\bf 6}(7), e22670.

\bibitem[Edmonds(1970)Edmonds]{Edmonds1970}
Edmonds, J. (1970).
\newblock Submodular functions, matroids, and certain polyhedra.
\newblock {\em Combinatorial Structures and Their Applications\/}, pages
  69--87.

\bibitem[Feige(1998)Feige]{Feige1998}
Feige, U. (1998).
\newblock A threshold of $\ln n$ for approximating set cover.
\newblock {\em JACM\/}, {\bf 45}(4), 634--652.

\bibitem[Fowler(1996)Fowler]{Fowler1996}
Fowler, D. (1996).
\newblock The binomial coefficient function.
\newblock {\em Am. Math. Monthly\/}, pages 1--17.

\bibitem[Fujishige(2005)Fujishige]{Fujishige2005}
Fujishige, S. (2005).
\newblock {\em Submodular functions and optimization\/}, volume~58.
\newblock Elsevier.

\bibitem[Ghiassian {\em et~al.}(2015)Ghiassian, Menche, and
  Barab{\'a}si]{Ghiassian2015}
Ghiassian, S.~D., Menche, J., and Barab{\'a}si, A.-L. (2015).
\newblock A {DIseAse} {MOdule} {Detection} ({DIAMOnD}) algorithm derived from a
  systematic analysis of connectivity patterns of disease proteins in the human
  interactome.
\newblock {\em PLoS Comput Biol\/}, {\bf 11}(4), e1004120.

\bibitem[Gon{\c c}alves {\em et~al.}(2012)Gon{\c c}alves {\em
  et~al.}]{Goncalves2012}
Gon{\c c}alves, J.~P. {\em et~al.} (2012).
\newblock Interactogeneous: Disease gene prioritization using heterogeneous
  networks and full topology scores.
\newblock {\em PLoS One\/}, {\bf 7}(11), e49634.

\bibitem[Gray {\em et~al.}(2015)Gray, Yates, Seal, Wright, and
  Bruford]{Gray2015}
Gray, K.~A., Yates, B., Seal, R.~L., Wright, M.~W., and Bruford, E.~A. (2015).
\newblock Genenames.org: the {HGNC} resources in 2015.
\newblock {\em NAR\/}, {\bf 43}(D1), D1079--D1085.

\bibitem[Greene {\em et~al.}(2015)Greene, Krishnan, Wong, {\em
  et~al.}]{Greene2015}
Greene, C.~S., Krishnan, A., Wong, {\em et~al.} (2015).
\newblock Understanding multicellular function and disease with human
  tissue-specific networks.
\newblock {\em Nature Genetics\/}.

\bibitem[Han {\em et~al.}(2013)Han {\em et~al.}]{Han2013}
Han, S. {\em et~al.} (2013).
\newblock Integrating {GWASs} and human protein interaction networks identifies
  a gene subnetwork underlying alcohol dependence.
\newblock {\em Am. J. Hum. Genet.}, {\bf 93}(6), 1027--1034.

\bibitem[Himmelstein and Baranzini(2015)Himmelstein and
  Baranzini]{Himmelstein2015}
Himmelstein, D.~S. and Baranzini, S.~E. (2015).
\newblock Heterogeneous network edge prediction: A data integration approach to
  prioritize disease-associated genes.
\newblock {\em PLoS Comput. Biol.}, {\bf 11}(7), e1004259.

\bibitem[Kibbe {\em et~al.}(2014)Kibbe {\em et~al.}]{Kibbe2014}
Kibbe, W.~A. {\em et~al.} (2014).
\newblock Disease {O}ntology 2015 update: an expanded and updated database of
  human diseases for linking biomedical knowledge through disease data.
\newblock {\em NAR\/}, page gku1011.

\bibitem[K{\"o}hler {\em et~al.}(2008)K{\"o}hler, Bauer, Horn, and
  Robinson]{Kohler2008}
K{\"o}hler, S., Bauer, S., Horn, D., and Robinson, P.~N. (2008).
\newblock Walking the interactome for prioritization of candidate disease
  genes.
\newblock {\em AJHG\/}, {\bf 82}(4), 949--958.

\bibitem[Krause and Guestrin(2011)Krause and Guestrin]{Krause2011}
Krause, A. and Guestrin, C. (2011).
\newblock Submodularity and its applications in optimized information
  gathering.
\newblock {\em TIST\/}, {\bf 2}(4), 32.

\bibitem[Lee {\em et~al.}(2004)Lee {\em et~al.}]{Lee2004}
Lee, I. {\em et~al.} (2004).
\newblock A probabilistic functional network of yeast genes.
\newblock {\em Science\/}, {\bf 306}(5701), 1555--1558.

\bibitem[Li and Patra(2010)Li and Patra]{Li2010}
Li, Y. and Patra, J.~C. (2010).
\newblock Genome-wide inferring gene--phenotype relationship by walking on the
  heterogeneous network.
\newblock {\em Bioinformatics\/}, {\bf 26}(9), 1219--1224.

\bibitem[Moreau and Tranchevent(2012)Moreau and Tranchevent]{Moreau2012}
Moreau, Y. and Tranchevent, L.-C. (2012).
\newblock Computational tools for prioritizing candidate genes: boosting
  disease gene discovery.
\newblock {\em Nat. Rev. Genet.}, {\bf 13}(8), 523--536.

\bibitem[Natarajan and Dhillon(2014)Natarajan and Dhillon]{Natarajan2014}
Natarajan, N. and Dhillon, I.~S. (2014).
\newblock Inductive matrix completion for predicting gene--disease
  associations.
\newblock {\em Bioinformatics\/}, {\bf 30}(12), i60--i68.

\bibitem[Navlakha and Kingsford(2010)Navlakha and Kingsford]{Navlakha2010}
Navlakha, S. and Kingsford, C. (2010).
\newblock The power of protein interaction networks for associating genes with
  diseases.
\newblock {\em Bioinformatics\/}, {\bf 26}(8), 1057--1063.

\bibitem[Nemhauser {\em et~al.}(1978)Nemhauser {\em et~al.}]{Nemhauser1978}
Nemhauser, G.~L. {\em et~al.} (1978).
\newblock An analysis of approximations for maximizing submodular set
  functions--{I}.
\newblock {\em Math. Programming\/}, {\bf 14}(1), 265--294.

\bibitem[Ritchie {\em et~al.}(2015)Ritchie {\em et~al.}]{Ritchie2015}
Ritchie, M.~D. {\em et~al.} (2015).
\newblock Methods of integrating data to uncover genotype-phenotype
  interactions.
\newblock {\em Nat. Rev. Genet.}, {\bf 16}(2), 85--97.

\bibitem[Ruffalo {\em et~al.}(2015)Ruffalo, Koyut{\"u}rk, and
  Sharan]{Ruffalo2015}
Ruffalo, M., Koyut{\"u}rk, M., and Sharan, R. (2015).
\newblock Network-based integration of disparate omic data to identify ``silent
  players'' in cancer.
\newblock {\em PLoS Comput. Biol.}, {\bf 11}(12).

\bibitem[Subramanian {\em et~al.}(2005)Subramanian {\em
  et~al.}]{Subramanian2005}
Subramanian, A. {\em et~al.} (2005).
\newblock Gene set enrichment analysis: a knowledge-based approach for
  interpreting genome-wide expression profiles.
\newblock {\em PNAS\/}, {\bf 102}(43), 15545--15550.

\bibitem[Sun {\em et~al.}(2011a)Sun {\em et~al.}]{Sun2011asonam}
Sun, Y. {\em et~al.} (2011a).
\newblock Co-author relationship prediction in heterogeneous bibliographic
  networks.
\newblock In {\em ASONAM\/}, pages 121--128.

\bibitem[Sun {\em et~al.}(2011b)Sun {\em et~al.}]{Sun2011vldb}
Sun, Y. {\em et~al.} (2011b).
\newblock Pathsim: Meta path-based top-k similarity search in heterogeneous
  information networks.
\newblock In {\em VLDB\/}.

\bibitem[Sun {\em et~al.}(2012)Sun {\em et~al.}]{Sun2012}
Sun, Y. {\em et~al.} (2012).
\newblock Integrating meta-path selection with user-guided object clustering in
  heterogeneous information networks.
\newblock In {\em KDD\/}, pages 1348--1356.

\bibitem[Ta{\c{s}}an {\em et~al.}(2015)Ta{\c{s}}an {\em et~al.}]{Tacsan2015}
Ta{\c{s}}an, M. {\em et~al.} (2015).
\newblock Selecting causal genes from genome-wide association studies via
  functionally coherent subnetworks.
\newblock {\em Nature Methods\/}, {\bf 12}(2), 154--159.

\bibitem[Vanunu {\em et~al.}(2010)Vanunu {\em et~al.}]{Vanunu2010}
Vanunu, O. {\em et~al.} (2010).
\newblock Associating genes and protein complexes with disease via network
  propagation.
\newblock {\em PLoS Comput. Biol.}, {\bf 6}(1), e1000641.

\bibitem[Wan {\em et~al.}(2015)Wan {\em et~al.}]{Wan2015}
Wan, C. {\em et~al.} (2015).
\newblock Classification with active learning and meta-paths in heterogeneous
  information networks.
\newblock In {\em CIKM\/}, pages 443--452.

\bibitem[Wang {\em et~al.}(2012)Wang {\em et~al.}]{Wang2012}
Wang, P.~I. {\em et~al.} (2012).
\newblock {RIDDLE:} reflective diffusion and local extension reveal functional
  associations for unannotated gene sets via proximity in a gene network.
\newblock {\em Genome Biology\/}, {\bf 13}(12), R125.

\bibitem[Warde-Farley {\em et~al.}(2010)Warde-Farley {\em et~al.}]{Warde2010}
Warde-Farley, D. {\em et~al.} (2010).
\newblock The {GeneMANIA} prediction server: biological network integration for
  gene prioritization and predicting gene function.
\newblock {\em NAR\/}, {\bf 38}(suppl 2), W214--W220.

\bibitem[Zhou {\em et~al.}(2014)Zhou {\em et~al.}]{Zhou2014}
Zhou, X. {\em et~al.} (2014).
\newblock Human symptoms--disease network.
\newblock {\em Nat. Commun.}, {\bf 5}, 4212.

\bibitem[Zitnik {\em et~al.}(2013)Zitnik {\em et~al.}]{Zitnik2013scirep}
Zitnik, M. {\em et~al.} (2013).
\newblock Discovering disease-disease associations by fusing systems-level
  molecular data.
\newblock {\em Scientific Reports\/}, {\bf 3}.

\bibitem[Zitnik {\em et~al.}(2015)Zitnik {\em et~al.}]{Zitnik2015ploscb}
Zitnik, M. {\em et~al.} (2015).
\newblock Gene prioritization by compressive data fusion and chaining.
\newblock {\em PLoS Comput. Biol.}, {\bf 11}(10), e1004552.

\bibitem[Zitnik and Zupan(2015)Zitnik and Zupan]{Zitnik2015tpami}
Zitnik, M. and Zupan, B. (2015).
\newblock Data fusion by matrix factorization.
\newblock {\em IEEE TPAMI\/}, {\bf 37}(1), 41--53.

\bibitem[Zitnik and Zupan(2016)Zitnik and Zupan]{Zitnik2016psb}
Zitnik, M. and Zupan, B. (2016).
\newblock Collective pairwise classification for multi-way analysis of disease
  and drug daata.
\newblock In {\em PSB\/}, volume~21, pages 81--92.

\end{thebibliography}

\end{document}